\documentclass{article}
\usepackage{ijcai22}

\usepackage{times}
\usepackage{soul}
\usepackage[utf8]{inputenc}
\usepackage{url}
\usepackage[colorlinks,linkcolor=blue]{hyperref}
\usepackage[small]{caption}
\usepackage{graphicx}
\usepackage{amsmath}
\usepackage{amsthm}
\usepackage{booktabs}
\usepackage{algorithm}
\usepackage{algorithmic}
\usepackage{dsfont}

\title{Goal-Conditioned Reinforcement Learning: Problems and Solutions}

\author{
Minghuan Liu\footnote{Equal contribution.}
\and
Menghui Zhu$^{*}$\and
Weinan Zhang\footnote{Corresponding author.}
\affiliations
Shanghai Jiao Tong University
\emails
\{minghuanliu,zerozmi7,wnzhang\}@sjtu.edu.cn
}

\newcommand{\citea}[1]{\citeauthor{#1} \shortcite{#1}}

\usepackage{amsmath,amsfonts,bm}

\def\eqref#1{equation~\ref{#1}}

\def\1{\bm{1}}

\DeclareMathAlphabet{\mathsfit}{\encodingdefault}{\sfdefault}{m}{sl}
\SetMathAlphabet{\mathsfit}{bold}{\encodingdefault}{\sfdefault}{bx}{n}

\DeclareMathOperator*{\argmax}{arg\,max}

\newtheorem{definition}{Definition}

\newcommand{\fig}[1]{Fig.~\ref{#1}}
\newcommand{\eq}[1]{Eq.~(\ref{#1})}
\newcommand{\tb}[1]{Tab.~\ref{#1}}
\newcommand{\se}[1]{Section~\ref{#1}}

\newcommand{\citet}[1]{\citeauthor{#1}~\shortcite{#1}}

\newcommand{\bbE}{\ensuremath{\mathbb{E}}} 
\newcommand{\bbR}{\ensuremath{\mathbb{R}}} 
\newcommand{\caA}{\ensuremath{\mathcal{A}}} 
\newcommand{\caS}{\ensuremath{\mathcal{S}}} 

\newcommand{\caG}{\ensuremath{\mathcal{G}}}

\usepackage{xcolor}

\newcommand{\revise}[1]{{#1}}

\begin{document}

\maketitle

\begin{abstract}
Goal-conditioned reinforcement learning (GCRL), related to a set of complex RL problems, trains an agent to achieve different goals under particular scenarios. Compared to the standard RL solutions that learn a policy solely depending on the states or observations, GCRL additionally requires the agent to make decisions according to different goals.
In this survey, we provide a comprehensive overview of the challenges and algorithms for GCRL. Firstly, we reveal the basic problems studied in this field. Then, we explain how goals are represented and present how existing solutions are designed from different points of view. Finally, we conclude and discuss potential future prospects that recent researches focus on.\footnote{We collect and summarize related algorithms and widely used benchmark environments in \url{https://github.com/apexrl/GCRL-Collection}.}
\end{abstract}

\section{Introduction}
One of the remarkable behaviors of human intelligence is that humans learn to reach one-by-one goals. Such ability allows generalizing their skills across different tasks while learning effectively. Building upon the recent successes of deep reinforcement learning (RL), which aims to induce a unique optimal solution for a specific task~\cite{sutton2018reinforcement}, researchers extend the problem towards tackling multiple goals simultaneously~\cite{plappert2018multi}, known as goal-conditioned RL, or goal-oriented RL, multi-goal RL, and we will refer as GCRL in this paper for brevity. 
\revise{
Technically, goal is a core data structure widely used in many related domains, for describing either achieving or maintaining some condition, e.g., goal reasoning~\cite{roberts2018special,niemueller2019goal} and robotics~\cite{rolf2013efficient,santucci2016grail}.
In this paper, we mainly focus on the usage of goals in RL literature. }
Different from standard RL, GCRL augments the observation with an additional goal that the agent is required to achieve when making a decision in an episode \cite{schaul2015universal,andrychowicz2017hindsight}. The reward function is known and usually defined as a binary bonus of reaching the goal. After training, the agent should be able to achieve arbitrary goals as the task specifies.
Such a problem raises additional challenges, for example, the simple but sparse reward function and the sample efficiency for multi-task learning.

From the task perspective, various solutions are designed for different categories of goals. 
A typical goal space is a sub-space of the state space, such as images or feature vectors, which is natural for describing the desired condition that the agents are required to meet. In addition, language, rewards, commands and intrinsic skills are also used as the representation of goals for different kinds of tasks.

From the learning perspective, researchers have proposed bags of tricks to alleviate the challenges in learning efficiency and generalization ability. Specifically, during a training loop, the agent first receives a desired goal from the environment or it should choose its own goal that is easier to achieve at the current stage~\cite{ren2019exploration,florensa2018automatic}. Then to pursue the goal, the agent interacts with the environment to generate trajectories~\cite{trott2019keeping}. The sampled trajectories are further post-processed for the training, for example, \citea{andrychowicz2017hindsight} chose to relabel the goals to increase the effectiveness of the samples. The final step in the loop is the optimization of the policy and value networks, where the goal-augmented extensions of the state and state-action value function are proposed as the foundation for solving the goal-conditioned policy~\cite{schaul2015universal}. 

This survey focuses on the algorithms designed for GCRL and is organized as follows:
\begin{enumerate}
    \vspace{-3pt}
    \item \textbf{Background and basic challenges}. We firstly formulate the basic problem in GCRL, and list all the challenges that should be addressed (\se{sec:formulation});
    \vspace{-3pt}
    \item \textbf{Goals representations}. We discuss how the goals are defined and represented in different task specifications (\se{sec:representation}); 
    \vspace{-3pt}
    \item \textbf{Solutions towards learning phases}. We categorize existing researches from the perspective of different phases, explain how the challenges of GCRL are tackled from different points of views (\se{sec:solutions});
    \vspace{-3pt}
    \item \textbf{Future prospects}. We summarize and present future prospects that are potential and expected to be studied in this field (\se{sec:conclusion}). 
\end{enumerate}


\section{Problems and Challenges}
\label{sec:formulation}

\subsection{Standard RL Problems as MDPs}
Standard RL tasks are usually modeled as Markov Decision Process (MDP) $\mathcal{M}$, denoted as a tuple $\langle \mathcal{S}, \mathcal{A}, \mathcal{T}, r, \gamma, \rho_0 \rangle$, where $\mathcal{S}$, $\mathcal{A}$, $\gamma$ and $\rho_0$ denote the state space, action space, discount factor and the distribution of initial states, respectively. $\mathcal{T}: \caS\times\caA\times\caS\rightarrow [0,1]$ is the dynamics transition function, 
and $r: \caS\times\caA\rightarrow \bbR$ is the reward function. Formally, the problem requires the agent to learn a decision-making policy $\pi:\caS\rightarrow\caA$ to maximize the expected cumulative return:
\begin{equation}
    J(\pi) = \mathbb{E}_{\substack{a_t \sim \pi(\cdot|s_t)\\s_{t+1} \sim \mathcal{T}(\cdot|s_t, a_t)}} \left [\sum_t \gamma^t r(s_t, a_t) \right ] ~.
\end{equation}


\subsection{Goal-Augmented MDPs}
\begin{figure}[htbp]
    \centering
    \includegraphics[width=.92\columnwidth]{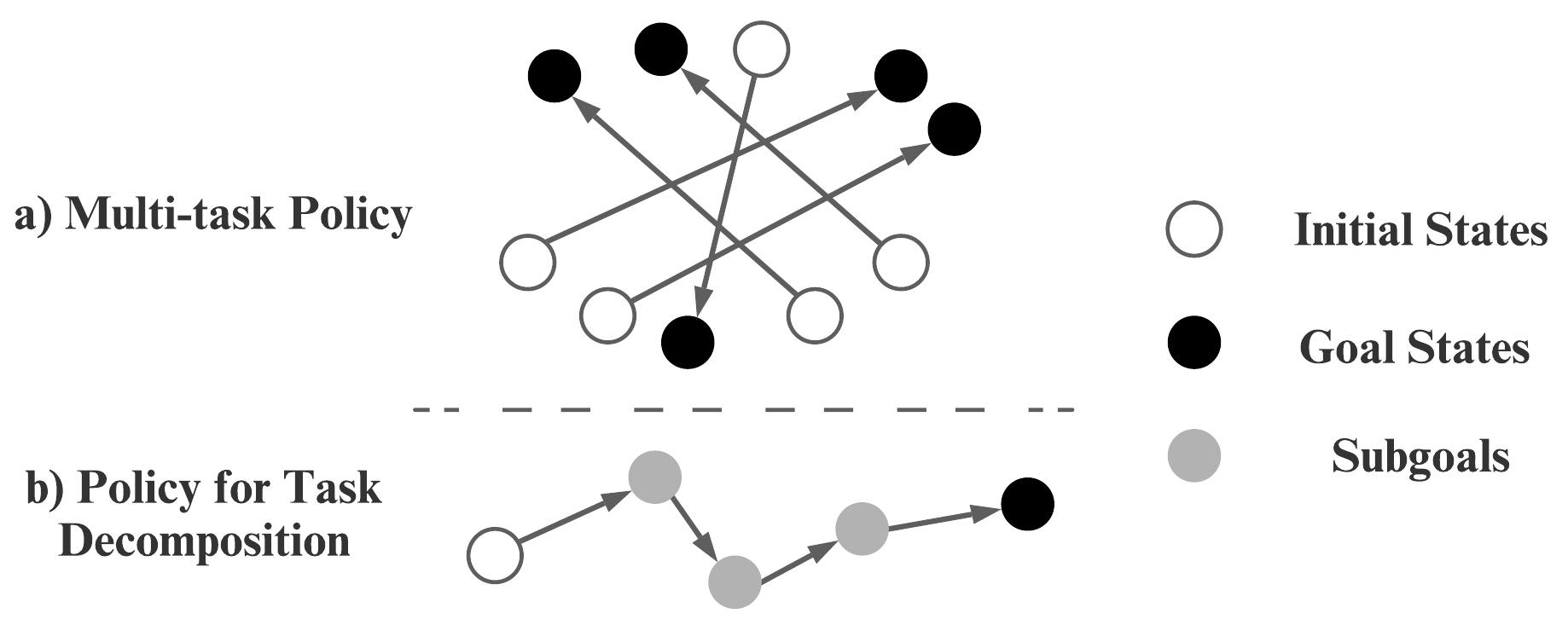}
    \caption{The mission of policy in complex RL problems. a) Learn to achieve multiple tasks with one single policy; b) Decompose the long-term, hard-reaching goals into easily obtained sub-goals.}
    \label{fig:complex-rl}
\end{figure}

Standard RL only requires the agent to finish one specific task defined by the reward function while GCRL focuses on a more general and complex scenario. As depicted in \fig{fig:complex-rl}, the agent in GCRL is required to either learn to master multiple tasks simultaneously, or decompose the long-term and hard-reaching goals into intermediate sub-goals while learning to reach them with a unified policy, or achieve both of them. \revise{For example, to navigate a robot or manipulate an object, the goals are typically defined as the target positions to reach. In \se{sec:representation}, we will further discuss the detailed representation of goals case-by-case.}

To tackle such a challenge, the formulation of GCRL augments the MDP with an extra tuple $\langle \caG, p_g, \phi \rangle$ as a goal-augmented MDP (GA-MDP), where $\caG$ denotes the space of goals describing the tasks, $p_g$ represents the desired goal distribution of the environment, and $\phi:\caS\rightarrow\caG$ is a tractable mapping function that maps the state to a specific goal. For many scenarios, $\caS$ and $\caG$ can be the same, in which case $\phi$ is an identity function. In GA-MDP, the reward function $r: \caS\times\caA\times\caG\rightarrow \bbR$ is defined with the goals, and therefore the objective of GCRL is to reach goal states via a goal-conditioned policy $\pi:\caS\times\caG\times\mathcal{A}\rightarrow[0,1]$ that maximizes the expectation of the cumulative return over the goal distribution:
\begin{equation}\label{eq:goal-object}
    J(\pi) = \mathbb{E}_{\substack{a_t \sim \pi(\cdot|s_t, g), g \sim p_g\\s_{t+1} \sim \mathcal{T}(\cdot|s_t, a_t)}} \left [\sum_t \gamma^t r(s_t, a_t, g) \right ] ~.
\end{equation}
To better understand the agent behaviors in a GA-MDP, some common definitions of goal used in previous works~\cite{pitis2020maximum,zhu2021mapgo}, are listed below.
\begin{definition}[Desired Goal]
A desired goal is a required task to solve which can either be provided by the environments or generated intrinsically. 
\end{definition}

\begin{definition}[Achieved Goal]
An achieved goal is the corresponding goal achieved by the agent at the current state.
\end{definition}

\begin{definition}[Behavioral Goal]
A behavioral goal represents the targeted task for sampling in a rollout episode.
\end{definition}
We note that the behavioral goal should be equivalent to the desired goal if there is no sub-goal replacement.

\subsection{Basic Challenges}
The mission of the agent as mentioned in \fig{fig:complex-rl} can be summarized as learning a multifunctional policy. In the sequel, the basic challenges are mainly the problem of generalization and sample efficiency.

\subsubsection{Generalization: Learning over Multiple Tasks}

In multi-task learning problems, denoting the loss function of the task $i$ as $L_{i}$, the weighted objective for learning $k$ tasks with a function $f$ simultaneously is defined as
\begin{equation}\label{eq:multi-task}
L(f) = \sum_{i=1}^{k}w_i \cdot L_i(f)~.
\end{equation}
This requires generalization and feature sharing across tasks that boost the learning efficiency of the function.
If we further regard the weight $w_i$ for each task $i$ as the occurring probability such that $i\sim p$ and $p(i)=w_i$,
we can further rewrite \eq{eq:multi-task} as an objective of expectation:
\begin{equation}\label{eq:multi-task-expectation}
L(f) = \bbE_{p}[L_i(f)]~.
\end{equation}
Notice that this generalizes the objective \eq{eq:goal-object} of GCRL. 
Correspondingly, the task variety $p$ is exactly the desired goal distribution $p_g$; and different goals $g\sim p_g$ implies different reward functions under the same environment. Since the agent is required to learn a unified policy to perform under multiple goals simultaneously, one of the main challenges in a complex RL problem can be naturally regarded as a multi-task learning problem towards generalization. And in result, we get to know that GCRL focuses on multi-task learning where the task variation comes only from the difference of the reward function induced by different desired goals under the same dynamics.

Fortunately, different goals in the same environment always possess similar properties (e.g., the dimension, the structure, and even the way to compute rewards are the same). This makes it possible for the goal-conditioned policy to share parameters across various tasks (i.e., goals $g$) to derive a generalized solution in GCRL. 


\subsubsection{Sample Efficiency: Towards Sparse Rewards}
\label{sec:sample-efficiency}
In GA-MDP, the reward function $r$ is typically a simple unshaped binary signal, indicating whether the task is completed, i.e.,
\begin{equation}\label{eq:goal-reward}
    r_g(s_t,a_t, g) = \mathds{1}({\text{the goal is reached}})~,
\end{equation}
where $\mathds{1}$ is the indicator function. The principle for completing the goal is usually a known criteria, for example, the distance between the achieved goal and the desired goal is less than a small number $\epsilon$~\cite{andrychowicz2017hindsight}: 
\begin{equation}\label{eq:indicator}
\mathds{1}({\text{the goal is reached}}) = \mathds{1}(\|\phi(s_{t+1})-g\| \leq \epsilon)~.
\end{equation}
Therefore, without shaping, the reward is rather sparse, so the agent can gain nothing from unsuccessful episodes. A straightforward idea to prevent the problem is to reshape the reward with a distance measure $d$ between the achieved goal and the desired goal without any domain knowledge~\cite{trott2019keeping}, i.e.,
\begin{equation}\label{eq:reshaping}
\tilde{r}_g(s_t, a_t, g) = -d(\phi(s_{t+1}), g)~.
\end{equation}
Such reshaped signals are so dense that the sparse problem can be alleviated in many cases. However, it may cause additional local optima challenges due to the structure of the environment. For instance, the reshaped reward prevents learning from a behavior that the agent must first increase the distance to the goal before finally reaching it~\cite{trott2019keeping}.

\section{What are Goals?}
\label{sec:representation}
In this section, we characterize the basic element in GCRL -- what exactly the goals are, and their typical representations. 
Indeed, there is no unique answer for the structure of goals, which highly depends on the task specifications. \revise{For research convenience, we further summarize the widely-used benchmark environments from previous works in \tb{tab:summary-envs}.}

\begin{figure}[htbp]
    \centering
    \includegraphics[width=.92\columnwidth]{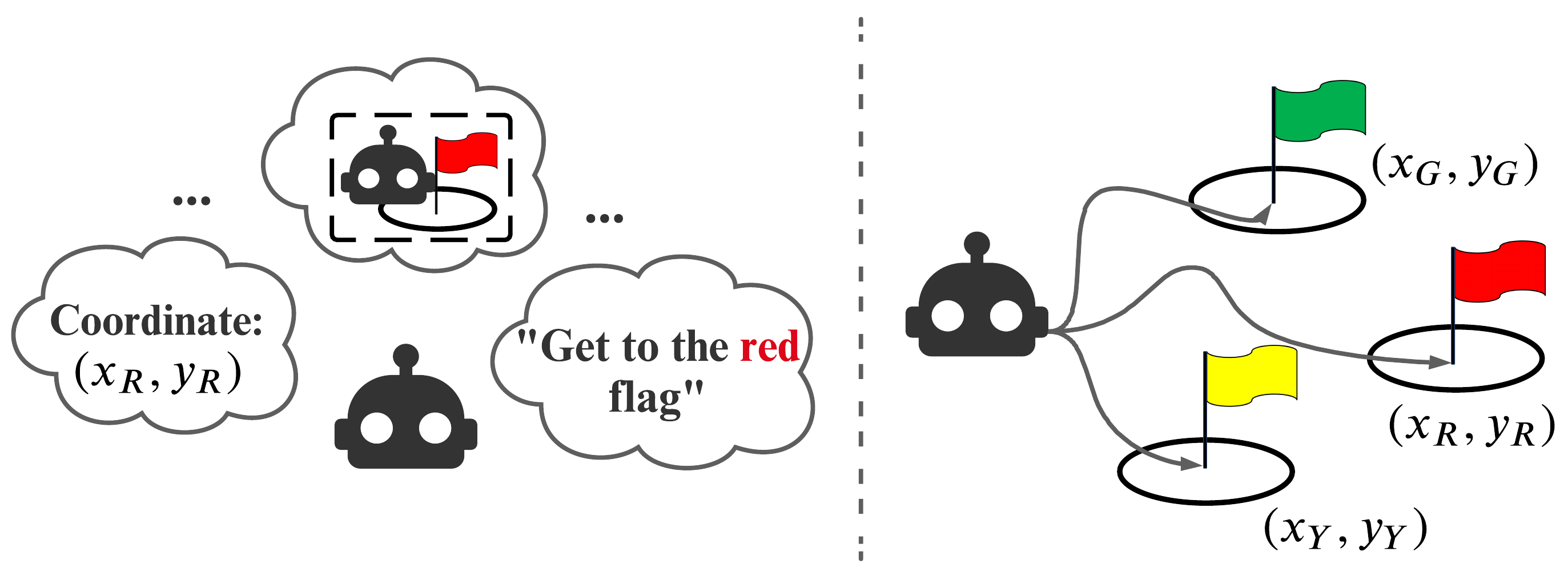}
    \caption{Typical representations of goals in GCRL: vectors, images and languages.}
    \label{fig:goal_types}
\end{figure}

\subsection{Desired Property and Feature Vectors}
\label{se:vec-goal}

Typically, goals are defined as desired properties or features of an agent. 
A task is therefore asking the agent to achieve some states that suffices specific features, usually represented as vectors. 
A wide variety of simulated robotics control tasks~\cite{plappert2018multi} are commonly used benchmarks for evaluating vector GCRL algorithms. For instance, asking the robotic arm to push, pick, or slide a block to a specific location, like \textit{FetchPush}, \textit{FetchPickAndPlace} and \textit{FetchSlide}. 
Additionally, a robotic hand is required to manipulate the block, or the pen to a particular position and rotation, e.g., \textit{HandManipulateBlock} and \textit{HandManipulatePen}. In such cases, the representation of the goal is a set of vectors of the desired location. 
On the other hand, setting the goal as a desired moving speed and ask the cheetah agent to run up at the speed (\textit{HalfCheetah})~\cite{zhu2021mapgo}; or a vector of the target position to reach, as is used in \textit{AntLocomotion}~\cite{florensa2018automatic}, \textit{Reacher}~\cite{tang2021hindsight}, and \textit{N-D Mass}~\cite{fang2019curriculum,pitis2020maximum}.


For tasks with vector goals, the goal space is usually a subspace of the state space. In other words, the mapping function $\phi$ is either an identical function such that matching a goal is exactly matching a target state, or a dimension reduction function that choosing specific dimensions from the state. 

\subsection{Image Goals}
Comparing to vectors describing simple properties and features, image is a more straightforward representation for complex tasks, especially in practical applications. Researchers used to evaluate their image algorithms on simulation environments, as in \citea{warde2018unsupervised,lee2020weakly,chane2021goal}, and also in real-world robotics applications~\cite{nair2018visual,nair2020goal}. 

Simulation environments include the robotics benchmarking tasks discussed in \se{se:vec-goal} with image state / goal space instead; moreover, DeepMind Lab~\cite{beattie2016deepmind}, Deepmind Control Suite (DMC)~\cite{tassa2020dmcontrol}, Atari games~\cite{brockman2016openai} and Minecraft~\cite{johnson2016malmo} are also popular testbeds. In these tasks, the goal is either the image of the target position to approach, or a imagined image of winning the game.
Beyond simulation, researchers also bring GCRL into real-world applications. 
For instance, \citea{nair2018visual,nair2020goal} first simulated a 7-dof Sawyer arm to reach goal positions, and then applied the agent to real world robotics control tasks with camera images.

To solve image tasks, the state and goal images can be directly fed into the policy and value networks and trained in an end-to-end manner, as in~\citea{chane2021goal,eysenbach2019search,trott2019keeping}. 
However, due to the high dimensionality of images, a more general way is to work on a compact vector representation, using an encoding model $E$ to map the image observation $o$ to a latent goal $g_z=E(o)$.
A common choice of the encoder is variational auto-encoder (VAE)~\cite{nair2018visual,khazatsky2021can}; or a context-conditioned VAE~\cite{nair2020contextual} for generalizing to different environments. The encoding can also be derived via training a weak supervised model for learning disentangled representation~\cite{lee2020weakly}; and even an end-to-end supervised imitation learning objective~\cite{srinivas2018universal}. In such tasks, the goal mapping function is an identical function since the state and goal are in the same image space.

\begin{table*}[t!]
\centering
\resizebox{0.97\textwidth}{!}{
\begin{tabular}{@{}llll@{}}
\toprule
\textbf{Environment / Task}     & \textbf{Benchmark Type}    & \textbf{Description}                                                                                                                                                                \\ \midrule
FetchReach             & Gym Robotics/Mujoco        & A robotic arm reaches a target position                                                                                                                                        \\  
FetchPush              & Gym Robotics/Mujoco        & A robotic arm pushes a block to a target position                                                                                                                                          \\ 
FetchPickAndPlace      & Gym Robotics/Mujoco        & A robotic arm first picks up the block then places it to a target position                                                                                                                 \\ 
FetchSlide             & Gym Robotics/Mujoco        & A robotic arm slides a buck to a goal position                                                                                                                                            \\ 
HandManipulateBlock    & Gym Robotics/Mujoco        & Orient a block using a robot hand                                                                                                                                          \\ 
HandManipulateEgg      & Gym Robotics/Mujoco        & Orient an egg using a robot hand                                                                                                                                           \\ 
HandManipulatePen      & Gym Robotics/Mujoco        & Orient a pen using a robot hand                                                                                                                                            \\ 
\midrule
HalfCheetah            & Mujoco                     & Make a 2D cheetah robot run and keep a specific speed                                                                                                                      \\ 
AntLocomotion / AntMaze & Mujoco                                 & Make a 3D four-legged robot walk to a target position                                                                                                                      \\ 
Sawyer                 & Mujoco                                 &  A sawyer robot reaches a target position                                                                                                              \\ 
Reacher                & Mujoco                     & A reacher robot reaches a target position                                                                                                             \\ 
N-D Mass               & Mujoco                                 & A ball reaches a target position                                                                                                                                        \\ 
DoorOpening            & Mujoco                                  &  A sawyer robot opens a door by latching onto the handle \\ 
Dy-Reaching            & Mujoco                                  & A robotic arm reaches a target position moving along a straight line \\ 
Dy-Circling            & Mujoco                                  & A robotic arm reaches a target position moving along a circle\\ 
Dy-Pushing             & Mujoco                                  & A robotic arm moves the box to a moving target position \\ 
Dy-Pouring             & Mujoco                                  & A robotic arm grips the can and pours the water into a cup                                                                                                      \\ 
\midrule
WaterMaze              & DeepMind Lab                            & Get to a target position described by an image                                                                                                                             \\ 
\midrule
Seaquest               & Atari                                   & Get to a target position described by an image                                                                                                                             \\ 
Montezuma’s Revenge    & Atari                                   & Get to a target position described by an image                                                                                                                             \\ 
\midrule
Reacher                & DeepMind Control Suit                   & A two-link planar reacher poses an aimed shape                                                                                                                   \\ 
Manipulator            & DeepMind Control Suit                   & A 3-DoF finger robot rotates the body into a target orientation                                                                                                  \\ 
Finger                 & DeepMind Control Suit                   & A planar manipulator brings an object in a target location                                                                                                       \\ 
\midrule
Place Object in Tray   & PyBullet                                & A robotic arm puts toys into a tray                                                                                                                                  \\ 
Opening Drawer         & PyBullet                                & A robotic arm opens the drawer                                                                                                                                       \\ 
PickandPlace           & PyBullet                                & A robotic arm first picks up the block then places it to a target position \\ 
\midrule
Five Object Manipulation          & CLEVR-Robot/Mujoco                                & A point mass agent arranges, orders and sorts 5 fixed objects \\ 
Diverse Object Manipulation          & CLEVR-Robot/Mujoco                                & A point mass agent orders objects in different shapes and colors \\ 
\midrule
Playground & Playground & A 2D hand moves and grasps/releases different objects\\
\midrule
VizDoom           & ViZDoom                                 & AI bots reach specific place in the first person shooter 3D
game Doom\\
\bottomrule
\end{tabular}
}
\caption{Summary of widely used benchmark environments. Note almost all environments support various kinds of goal structures, i.e., vector, image, language, etc. More detailed information can be referred to the \href{https://github.com/apexrl/GCRL-Collection}{repository} we maintained.}
\label{tab:summary-envs}
\end{table*}

\subsection{Language Goals}
Natural languages are also used to represent the goals. In fact, language-conditioned RL is a wide research direction, which combines the techniques of GCRL and the development of neural language processing (NLP). Since our work focus on the essential of GCRL, we will only cover how language forms the goals, and readers may refer to \citea{luketina2019survey} for a detailed review of language usage in RL.

The advantage of language goals, i.e., sequences of discrete symbols, is that they can easily be understood by human. 
In most cases, the goal is an instruction sentence containing explicit verbs and objects, for instance, ``go to blue torch"~\cite{chan2019actrce};
or question words ``there is a red ball; are there any matte cyan sphere right of it?"~\cite{jiang2019language}; and simply, a predefined set of semantic predicates, like ``close", ``above"~\cite{akakzia2020grounding}. To train RL algorithms, those words must be either translated into embeddings via pre-training models~\cite{chan2019actrce}, or taken into a recurrent neural network (RNN) as learning the embedding jointly with the RL agent~\cite{colas2020language}. If the language space is constrained such that the discrete predicates or objects are limited, a simple semantic mapping function can be learned to translate the handcrafted encodings into vectors~\cite{akakzia2020grounding}. In this case, the goal mapping is a function from the state space to the embedding space.


\subsection{Other Forms of Goals}
\label{sec:other-form}
Not limited in forms above, researchers have used the conditioned policy on various goal types, which may lead to different unresolved problems. 

\paragraph{Expected return as goals.} \citea{kumar2019reward} proposed to learn a return-conditioned policy $\pi(a|s,R)$ where $R$ is an estimated return of the whole trajectory. 
The basic idea is to reuse sub-optimal trajectories for training as these trajectories are optimal supervision for a policy of certain quality, and thus the return $R$ can be seen as a desired goal to reach. Specifically, $R$ is modeled as a dynamic distribution as the learning proceeds. 
Recently, \citea{chen2021decision} proposed decision transformer (DT) to solve offline RL problems, which treats RL as a sequence modeling problem by generating actions from offline datasets. In detail, the trajectories are represented as a sequence of $(\widehat R_1, s_1, a_1, ..., \widehat R_T, s_T, a_T)$, where $\widehat R_t=\sum_{t'=t}^{T} r_{t'}$ is the target reward that requires the agent to reach, and resembles the desired goal in GCRL.

\paragraph{Upside-down RL: commands as goals.} Concurrently, \citea{schmidhuber2019reinforcement} and \citea{srivastava2019training} also exploited reward-form goals on standard RL problems. Different from the works discussed above, they regard the goals as a set of commands, such as asking the agent to reach a desired return $R$ in a desired horizon $H$, leading to a conditioned policy $\pi(a|s, R, H)$. These works do not adopt any training techniques from RL literature, but learn the policy from replay buffer in a supervised style instead.

\begin{figure*}[!htbp]
\centering
\includegraphics[width=0.92\linewidth]{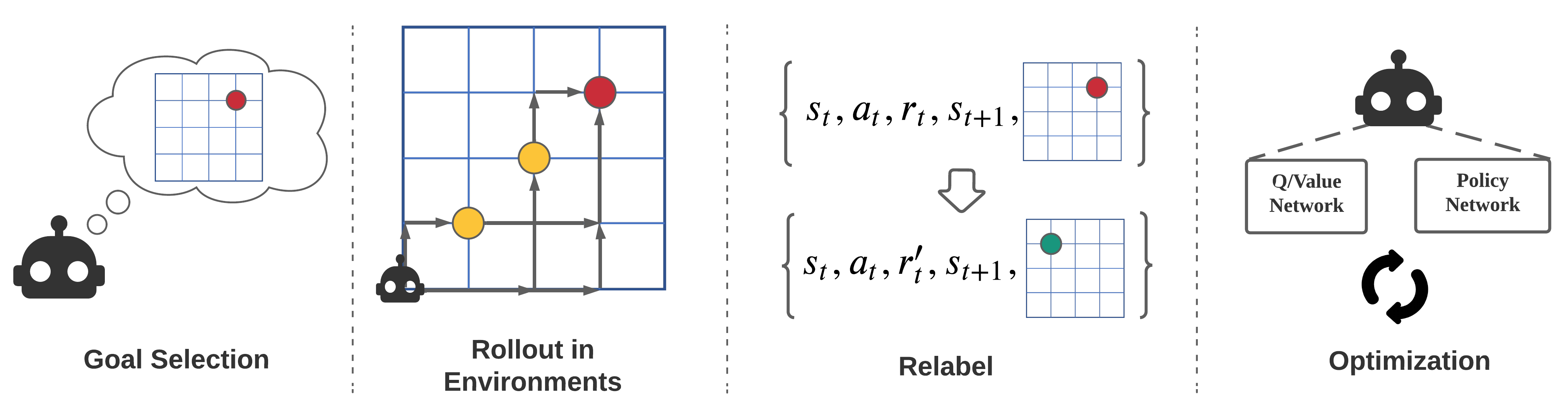}
\caption{Typical phases in a GCRL algorithm. The agent is provided (or selects by itself) a behavior goal as the target, then interacts with the environments and collects experiences. Before optimizing the policy, the historical data can be relabeled by changing the desired goals.}
\label{fig:goal_process}
\end{figure*}

\section{Algorithms to GCRL}
\label{sec:solutions}
In this section, we discuss the solutions for GCRL tasks from the major perspectives of algorithms, including optimization, sub-goal selection and relabeling, as shown in \fig{fig:goal_process}, \revise{which is also summarized in \tb{tab:summary-works}.}

\subsection{Optimization}
The essential difference between GCRL and traditional RL algorithms on the stage of optimization is learning a conditioned policy towards a set of goals. Therefore, both universal and specific solutions are proposed for resolving GCRL.

\paragraph{Universal value function.} To solve GCRL,  \citea{schaul2015universal} first proposed the universal value function approximation (UVFA), which provides a straightforward way to characterize Goal-Augmented MDP (GA-MDP) by generalizing the standard value function as a goal-conditioned value function, and is defined as
\begin{equation}
    V^{\pi}(s_t, g) = \mathbb{E}_{\substack{a_t \sim \pi(\cdot|s_t, a_t)\\s_{t+1} \sim \mathcal{T}(\cdot|s_t, a_t)}} \Big[ \sum_{t} \gamma^t r(s_t,a_t,g) \Big]~.
\end{equation}
UVFA is widely applied in the following research works serving as a basic framework.

\paragraph{Reward shaping.} One of the basic challenges of GCRL is the extremely sparse reward, so a corresponding solution should be reward shaping. In the previous context, we have explained that a lazy distance-based shaping method can lead to additional local optima. To alleviate the problem, \citea{trott2019keeping} proposed a sibling rivalry (SR) mechanism, which samples two sibling rollouts simultaneously, uses each others' achieved goals as anti-goals $\bar g$ and reward an anti-goal bonus $d(s, \bar g)$ to improve exploring, where $d$ is the distance metric, such as L2-norm. In addition, many works proposed different distance measures for evaluating the progress of achieving goals. For instance, \citea{hartikainen2019dynamical} calculated the average number of actions required to taken to reach the goal and used it as the penalty on reward; \citea{wu2018laplacian} learned the Laplacian matrix then used its eigenvectors as the representation of the states and goals, and the reward function is defined as the Euclidean distance between the state and goal vectors; \citea{durugkar2021adversarial} defined the target distribution in GCRL as a Dirac measure at the goal state and estimated Wasserstein-1 distance of the state visitation distribution to the target distribution as the reward bonus. In order to encourage exploration, \citea{mccarthy2021imaginary} trained an ensemble of forward models and adopted the variance between the ensemble predictions of the next state as the intrinsic reward. 

\paragraph{Self-imitation learning.} Although reaching a specific goal state is agnostic to the agent at the beginning of training, we can treat any trajectory as a successful trial for reaching its final state. Drawn intuition from such an observation, \citea{ghosh2019learning} proposed a supervised algorithm named goal-conditioned supervised learning (GCSL), which learns the policy by maximizing the likelihood of the actions for a reached goal within a number of steps:
\begin{equation}
    J(\pi) = \mathbb{E}_{\mathcal{D}} \left [ \log \pi(a|s,g,h)\right ] ~,
\end{equation}
where $\mathcal{D}$ is the replay buffer and $h$ represents the left steps (horizon) to reach the goal. 
Similar to the idea of GCSL, \citea{nair2018overcoming} filtered the actions from demonstrations by Q values and added a supervised auxiliary loss on the RL objective to improve the learning efficiency.

\paragraph{Planning and model-based RL.}
In RL tasks, agents must interact with the environment to collect experience for policy learning, which results in considerable sampling costs. However, as humans have the ability to infer future outcomes in mind, agents can also keep an environment dynamics model for planning or imagining with less sampling in the environment. Formally, the probabilistic parameterized dynamics model $M_{\theta}$ can be trained from experience data $\mathcal{D}$ by supervised objectives, e.g., by maximum likelihood estimation:
\begin{equation}
\begin{aligned}
    \argmax_{\theta} \bbE_{\{s_t, a_t, s_{t+1}\} \sim \mathcal{D}} \left [\log Pr(s_{t+1}|s_t, a_t; \theta)\right ]~.
\end{aligned}
\end{equation}
Generally speaking, algorithms utilizing a dynamics model are realized as model-based methods. The typical usage of the model contains planning and generating more interaction data for training to improve the sample efficiency.
For example, \citea{nair2020goal} tried to plan to reach the goal states in the latent space for image-based inputs; they trained an encoder-decoder model and a goal-conditioned dynamics model on the latent space to learn the state representation function.
\citea{charlesworth2020plangan} utilized a dynamics model for predicting consecutive states to regularize the consistency of the trajectories generated by a GAN model. 
With an ensemble of well-trained GANs, the agent generate trajectories and choose the next actions from the trajectory that spends less time achieving the goal. Besides, 
to decrease the sampling cost, \citea{zhu2021mapgo} brought model-based optimization techniques into GCRL while also planning in the learned model to infer future goals for relabeling;
\citea{zhang2021world} learned a representation model from the goal space to a compact latent space in a self-supervised manner, and tried to find a sparse set of latent landmarks scattered across goal space in terms of reachability by clustering and planning with the landmarks;
\citea{mendonca2021discovering} took an ensemble of dynamics model for training an explorer to discover unseen surprising states by defining exploration bonus as the variance of the ensemble predictions, and they further train the policy to reach randomly sampled goals from the replay buffer.


\subsection{Sub-goal Selection / Generation} 
In most environments, the desired goal asked by the task for the agent is a long-term target and hard to achieve. Nevertheless, building on the insight that humans tend to split the long-term target into the short-term goals that are easier to reach, researchers have proposed to generate or select sub-goals to help the agent reach the target goals. Formally, collecting experience following a replaced desired goal is changing the goal distribution of the environment from $p_g$ into $f$, and the optimization in \eq{eq:goal-object} is now rewritten as
\begin{equation}\label{eq:goal-object-subgoal}
    J(\pi) = \mathbb{E}_{\substack{a_t \sim \pi(\cdot|s_t, g), g \sim f\\s_{t+1} \sim \mathcal{T}(\cdot|s_t, a_t)}} \left [\sum_t \gamma^t r(s_t, a_t, g) \right ] ~,
\end{equation}
where $f$ can either be specific rules that select $g$ from past experience, or a learned function depending on the principle.
\paragraph{Intermediate difficulty.}
Accurately evaluating the difficulty to reach a goal helps determine what to learn within the capability of the agent. An appropriate goal should be just beyond the scope of the agent in order to improve learning efficiency. Accordingly, the intermediate difficulty can be used as a critic principle for choosing sub-goals. 
For instance, \citea{florensa2018automatic} scored the difficulty with the received reward, which is used to label the goals as positive/negative samples; then, they trained a GAN model to generate goals with appropriate difficulty scores.
\citea{campero2020learning} proposed a teacher-student framework, where the teacher serves as a goal producer that tries to propose increasingly challenging goals. The teacher learns from the student's performance on the produced goal, while the student learns from both extrinsic rewards from the environment and intrinsic rewards from the teacher.
Moreover, \citea{zhang2020automatic} treated the epistemic uncertainty of the Q-function as a measurement of the difficulty for reached goals, which is further used for constructing a distribution to sample goals. When the uncertainty is high, the goal should be lied at the knowledge frontier of the policy, and is thus at an appropriate difficulty to learn. 

\paragraph{Exploration awareness.} Notice that in GCRL, the essential challenge making desired goals hard to reach is the sparse reward function which limits the learning efficiency of the agent. Therefore, we should boost its exploration ability to cover unseen goals and states as many as possible, so that the agent can learn far more from experience replay.
To this end, in GCRL, many researchers propose to sample with the sub-goals that lead to a better exploring ability.
\citea{pitis2020maximum} proposed to sample the goals which could maximize the entropy of the historically achieved goals, and in result, enhance the coverage of the goal space;
\citea{pong2019skew} trained a self-supervised generative model to generate behavioral goals that are skewed to be approximately uniform over achieved goals from past experience, which is shown as equivalent to maximizing the entropy of the desired goal distribution.
Likewise, to provide with reachable and diverse goals, \citea{warde2018unsupervised} and \citea{nair2018visual} both chose to sample past achieved goals uniformly from experience replay as the behavioral goals, while \citea{warde2018unsupervised} sampled goals from a diversified replay buffer in which the distance between the stored goals are held as far as possible, and \citea{nair2018visual} generate goals by fitting a generative model for historical achieved goals.
Furthermore, \citea{ren2019exploration} generated a set of hindsight goals for exploration by formulating a Wasserstein Barycenter problem. The generated sub-goals serve as an implicit curriculum that is effective to learn using the current value function.
\citea{hartikainen2019dynamical} chose to learn a distance function and take the goals from the replay buffer which have maximum distance from the initial state for exploration efficiency.

\paragraph{Searching from experience.}
Looking back upon past experience, both humans and agents can benefit since the history contains the way-points leading us to reach some goals. 
Motivated by such an idea, \citea{lai2020hindsight} chose $k$ critical middle point goals from a sampled episode, and learned an RNN to generate the sequence given the starting and the ultimate point; thereafter, they apply the generated middle points as a sequence of desired goals when interacting with the environment. On the other hand, \citea{eysenbach2019search} built a graph on the replay buffer, taking the states as nodes, and distance from the start state to the goal state as the edge weights. In their work, they utilize the intuition that the value of states can approximate the distance to goals under the binary reward function (\eq{eq:goal-reward}); then, the proposed algorithm runs a graph search to find the sequence of way-points to achieve the target and takes actions by a learned policy conditioned on the way-points one after another.

\paragraph{Model-based planning.} 
Human has the ability to plan a long-term goal by decomposing it into several short-term segments. To this end, \citea{nair2019hierarchical} trained a dynamics model for planning a sequence of sub-goals from the initial state to the final target in image-based tasks. Particularly, they performed visual model-predictive control (MPC) to measure the predicted cost of the sub-goal segments and sought for the minimal-cost one as the current goal to reach. When the sub-goals were planned, they simply used visual MPC to plan to reach sub-goals one by one. 

\paragraph{Learn from experts.} The expert demonstration can provide useful guidance for reaching a hard goal, since it directly contains the way-points for decomposing such a target into smaller sub-goals. 
Such an intuition motivates the work of \citea{paul2019learning}, which made use of the order of sub-goal indices along the expert trajectory as the evidence of difficulty, and learned to predict the sub-goals in the equipartitions from the expert trajectory conditioned on the given state.

\begin{table*}[t!]
\centering
\resizebox{0.93 \textwidth}{!}{
    \begin{tabular}{@{}lcccc@{}}
    \toprule
        \textbf{Approach} & \textbf{Resolved Goal Type} & \textbf{Sub-Goal} & \textbf{Relabel} & \textbf{Optimization} \\ 
        \midrule
        UVFA \cite{schaul2015universal} & Vector/Image & ~ & ~ & \checkmark \\ 
        HER \cite{andrychowicz2017hindsight} & Vector/Image & ~ & \checkmark & ~ \\ 
        Laplacian Representation Learning \cite{wu2018laplacian} & Vector/Image & ~ & ~ & \checkmark \\ 
        SR \cite{trott2019keeping} & Vector/Image & ~ & ~ & \checkmark \\ 
        DDL \cite{hartikainen2019dynamical} & Vector/Image & \checkmark & ~ & \checkmark \\ 
        SoRB \cite{eysenbach2019search} & Vector/Image & \checkmark & ~ & ~ \\ 
        Skew-Fit \cite{pong2019skew} & Vector/Image & \checkmark & ~ & ~ \\
        \midrule
        HER with Demonstrations \cite{nair2018overcoming} & Vector & ~ & ~ & \checkmark \\ 
        GoalGAN \cite{florensa2018automatic} & Vector & \checkmark & ~ & ~ \\ 
        Sub-goal Discovery \cite{paul2019learning} & Vector & \checkmark & ~ & ~ \\ 
        HGG \cite{ren2019exploration} & Vector & \checkmark & ~ & ~ \\ 
        CHER \cite{fang2019curriculum} & Vector & ~ & \checkmark & ~ \\ 
        G-HER \cite{bai2019guided} & Vector & ~ & \checkmark & ~ \\
        Hindsight Planner \cite{lai2020hindsight} & Vector & \checkmark & ~ & ~ \\ 
        GDP \cite{kuang2020goal} & Vector & ~ & \checkmark & ~ \\ 
        VDS \cite{zhang2020automatic} & Vector & \checkmark & ~ & ~ \\ 
        MEGA \cite{pitis2020maximum} & Vector & \checkmark & \checkmark & ~ \\ 
        PlanGAN \cite{charlesworth2020plangan} & Vector & ~ & ~ & \checkmark \\ 
        AIM \cite{durugkar2021adversarial} & Vector & ~ & ~ & \checkmark \\ 
        I-HER \cite{mccarthy2021imaginary} & Vector & ~ & ~ & \checkmark \\ 
        GCSL \cite{ghosh2019learning} & Vector & ~ & ~ & \checkmark \\
        MapGo \cite{zhu2021mapgo} & Vector & ~ & \checkmark & \checkmark \\ 
        L3P \cite{zhang2021world} & Vector & ~ & ~ & \checkmark \\
        \midrule
        RIG \cite{nair2018visual} & Image & \checkmark & ~ & ~ \\ 
        DISCERN \cite{warde2018unsupervised} & Image & \checkmark & ~ & ~ \\ 
        HVF \cite{nair2019hierarchical} & Image & \checkmark & ~ & ~ \\ 

        AMIGo \cite{campero2020learning} & Image & \checkmark & ~ & ~ \\ 
        GAP \cite{nair2020goal} & Image & ~ & ~ & \checkmark \\ 
        LEXA \cite{mendonca2021discovering} & Image & ~ & ~ & \checkmark \\ 
        \bottomrule
    \end{tabular}
}
\caption{Summary of the algorithms mentioned in \se{sec:solutions}.}
\label{tab:summary-works}
\end{table*}

\subsection{Relabeling} 
Similar to sub-goal selection, the relabeling step also replaces the initial desired goal provided by the task to enhance the learning efficiency. However, relabeling focuses on replacing the history data in the experience buffer before training the agent, instead of changing the distribution for sampling experience. Formally, it requires a relabel function $h$ to replace the goal of the samples tuple $(s_t,a_t,g_t,r_t)$ from the replay buffer $\mathcal{B}$ and re-compute the reward:
\begin{equation}\label{eq:goal-object-relabel}
     (s_t,a_t,g_t,r_t)\leftarrow (s_t,a_t,h(\cdot),r(s_t,a_t,h(\cdot)))~,
\end{equation}
where $h$ can be designed as taking any particular features into getting new goals.

\paragraph{Hindsight relabeling.}
The relabeling strategy can be traced back to the famous Hindsight Experience Replay (HER)~\cite{andrychowicz2017hindsight}, which is motivated by the fact that human learns from failure experience. HER relabels the desired goals in the buffer with achieved goals in the same trajectories and alleviates the sparse reward signal problem. It shows that off-policy algorithms can reuse the data in the replay buffer by relabeling goals in the previous trajectory. 
HER is widely used as the backbone in many related works~\cite{ren2019exploration,pitis2020maximum,durugkar2021adversarial}, and researchers also take efforts on extending the idea of HER on designing advanced relabeling strategies. As an improvement, \citea{fang2019curriculum} proposed CHER, a curriculum relabeling method which adaptively selects the relabeled goals from failed experiences. The curriculum is evaluated according to the likeness to desired goals and the diversity of goals.
To increase the variety of relabeled goals and not restricted in the same trajectory, \citea{pitis2020maximum} generalized HER by additionally relabeling goals that are randomly sampled from buffers of desired goals, achieved goals and behavioral goals, which allows the agent to learn different skills at a time.
To improve the sample efficiency by exploring on rare visited experiences, \citea{kuang2020goal} learned a density model for achieved goals and relabeled them by prioritizing the rarely seen ones using the density model.

Recently, \citea{eysenbach2020rewriting} revealed the relationship between HER and inverse RL which studies how to recover the reward function on a set of offline expert trajectories. Specifically, hindsight relabeling can be seen as a special case of maximum-entropy inverse RL with a pre-defined reward function, and thus inverse RL method is used for relabeling goals in their work.

\paragraph{Relabeling by learning.}
Selecting the relabeling goals from past experiences limit the diversity of relabeled goals. To alleviate such a problem and allows relabeling unseen goals, \citea{bai2019guided} trained a conditional generative RNN by explicitly modeling the relationship between the policy performance and the achieved goals, which is further used for generating relabeled goals according to the averaged return obtained by the current policy.


\paragraph{Foresight.}
Motivated by the fact that humans can learn from failure but are also capable of planning for the future based on the current situations, \citea{zhu2021mapgo} proposed to learn and utilize a dynamics model to generate virtual trajectories for relabeling. Foresight goal relabeling can also prevent labeling homogeneous goals limited in historical data, which plans new goals that the current policy can achieve.





\section{Summary and Future Prospects}
\label{sec:conclusion}
This paper provides a brief review of the development of goal-conditioned reinforcement learning (GCRL). We show the formulation in the beginning, explain the problem and summarize the basic challenges of in GCRL literature. Further, we investigate the critical element -- goal in previous works and present its main types. In the later part, we discuss key ways for solving GCRL from the perspective of optimization, sub-goal generation and relabeling, respectively. 

Although we try to cover most of the topic, GCRL formulation and its variant are not limited in the discussed material. \revise{Interested readers can also refer to another survey paper~\cite{colas2020intrinsically} that concluded GCRL from a different point of view.}
As is known to all, RL is hard for large-scale applications, due to its thorniest limitation of learning from a single reward function and the bottleneck of the online interaction cost. Therefore, we note four potential prospects for GCRL, namely learning totally intrinsic skills, learning from offline datasets, and large decision models.

\paragraph{Learning totally intrinsic skills.}
The formulation of GCRL provides the considerable potential to incorporate self-supervised training to improve the model ability instead of learning from a simple reward signal. Recently, there are more research works that break the circle of reaching a meaningful goal state (we have discussed some examples like reward and commands in \se{sec:other-form}) and ask the agents to explore all possible goals through itself, as a trend of developing totally intrinsic skill learning~\cite{eysenbach2018diversity,sharma2019dynamics}. Specifically, a goal-conditioned policy can be only with intrinsic rewards targeted to maximizing the mutual information between the skill and the behaviors.
\paragraph{Learning from offline datasets.}
Since the online data collection in many realistic applications is always infeasible, many researches pay attention to offline RL methods, which shows potential to learning sequential decision policies solely from static datasets without interacting with the environment. This inspires taking advantage of GCRL for learning generalized goal-conditioned decision models from the offline dataset, like in \citea{tian2020model} and \citea{chebotar2021actionable}.
\revise{
\paragraph{Potential in large decision models.}
An appealing direction for utilizing GCRL is to develop general pre-trained goal-conditioned control policies. Combined with other pre-trained models, such a large decision model can be used to command the robots to handle a variety of tasks with kinds of goal elements, such as a sentence of text description~\cite{ahn2022can}, a goal image of the order of object placement~\cite{chebotar2021actionable}, a goal image of the targeted area in the open world~\cite{shah2021rapid}, etc.}

\paragraph{Other prospects.}
Apart from directions mentioned above, we note that there are more intriguing future directions worth explored in the formulation of GCRL, such as time-extended goals~\cite{forestier2016modular,brafman2018ltlf}, constraint-based goals~\cite{colas2021epidemioptim}, composition of goals~\cite{chitnis2020glib} etc. 





\section*{Acknowledgements}
The authors are supported by ``New Generation of AI 2030'' Major Project (2018AAA0100900), Shanghai Municipal Science and Technology Major Project (2021SHZDZX0102), NSFC (62076161).
Minghuan Liu is also supported by Wu Wen Jun Honorary Doctoral Scholarship, AI Institute, SJTU. We thank Mingcheng Chen, Xihuai Wang and Zhengyu Yang for helpful discussions.

\bibliographystyle{named}
\bibliography{ijcai22}

\end{document}